\documentclass{ifacconf}

\usepackage{graphicx}      
\usepackage{natbib}
\usepackage{import}

\usepackage{amsmath,amsfonts} 
\usepackage{graphicx,xcolor}
\usepackage{tabularx,booktabs} 
\usepackage{import}
\usepackage{times,xspace} 
\usepackage{mathtools}
\usepackage{bm}
\usepackage{romannum}
\usepackage{siunitx}
\usepackage{xargs}
\usepackage{balance}
\usepackage{url}
\usepackage{leftidx}
\usepackage[font={footnotesize}]{subcaption}
\usepackage[font={footnotesize}]{caption}
\usepackage{tcolorbox}
\usepackage{algorithm}
\usepackage{setspace}
\usepackage{algpseudocode} 
\usepackage{adjustbox}
\usepackage{enumitem}
\usepackage{makecell}
\begin{document}
\begin{frontmatter}

\title{Branch-Stochastic Model Predictive Control for Motion Planning under Multi-Modal Uncertainty with Scenario Clustering} 

\thanks[footnoteinfo]{\copyright\ 2026 the authors. This work has been accepted to IFAC for publication under a Creative Commons Licence CC-BY-NC-ND}

\author[First]{Zekun Xing} 
\author[First]{Ramkrishna Chaudhari} 
\author[First]{Marion Leibold}
\author[First]{Dirk Wollherr}
\author[First]{Martin Buss}

\address[First]{Chair of Automatic Control Engineering, Technical University of
Munich, Arcisstr. 21, 80333 Munich, Germany (e-mail: \{zekun.xing, ramkrishna.chaudhari, marion.leibold, dirk.wollherr, mb\}@tum.de).}

\begin{abstract}                
Motion planning for autonomous driving must account for multi-modal uncertainty in both the intentions and trajectories of surrounding vehicles. Handling uncertainty in a worst-case manner guarantees robustness but often leads to excessive conservatism. Stochastic Model Predictive Control (SMPC) reduces trajectory-level conservatism through chance constraints, yet remains conservative with respect to intention uncertainty since constraints must hold across all intentions. We present a novel combination of SMPC and the branching structure, enabling the planner to generate distinct trajectories for different possible intentions while maintaining safety under trajectory uncertainty. A novel scenario clustering is proposed to merge prediction scenarios based on high-level decision similarity, thereby ensuring real-time tractability. Furthermore, an adaptive branching-time computation postpones commitment to separate plans until intention uncertainty is sufficiently reduced. Simulation studies in challenging highway scenarios demonstrate that the proposed method improves safety, reduces conservatism, and achieves real-time computational performance.
\end{abstract}


\end{frontmatter}
\section{Introduction}
\label{sec:introduction}
Autonomous vehicles (AVs) must operate safely and efficiently in dynamic and uncertain traffic environments where the future behaviors of surrounding vehicles (SVs) are interactive and multi-modal \citep{wang2022social}. 
Multi-modality means for us that uncertainties of SVs arise from both intention-level and trajectory-level factors \citep{benciolini2023non, zhou2023interaction}. 
On the one hand, SVs interact with nearby vehicles, including the AV, meaning that their future behaviors depend not only on their own objectives but also on how the nearby vehicles act. This interaction results in multiple plausible future behaviors and introduces uncertainty at the intention level.
On the other hand, even when an intention is inferred, the trajectory execution of that intention is not perfectly predictable, since the driving style and dynamics of SVs are only partially observable. 
Modern learning-based \citep{salzmann2020trajectron++, shi2022motion} and interaction-aware predictors \citep{lefkopoulos2020interaction, liu2022interaction} can capture interactions and multi-modal uncertainty of SVs, and generate accurate probabilistic multi-modal trajectory predictions.
However, leveraging such interactive and multi-modal predictions for real-time, safe motion planning remains challenging. The prediction complexity scales with the number of SVs, and the planner must reason over many uncertain future behaviors of SVs simultaneously. Only enumerating all plausible behaviors without structure can make the planning problem intractable and overly conservative.

Among existing motion-planning approaches, optimization-based methods with Model Predictive Control (MPC) are particularly well-suited for AVs, as they explicitly incorporate vehicle dynamics, traffic rules, safety constraints, and multi-objective trade-offs \citep{levinson2011towards}. 
To handle uncertainty, numerous variations of MPC have been proposed and implemented in the field of autonomous driving. Robust MPC (RMPC) \citep{massera2017safe} enforces constraint satisfaction for all possible uncertainty realizations, ensuring safety but often yielding overly conservative trajectories, particularly when uncertainty sets are large or hard to characterize.
Scenario MPC (SCMPC) approximates stochastic uncertainty using a set of sampled scenarios and optimizes a trajectory that satisfies the constraints for all samples \citep{cesari2017scenario, kensbock2023scenario}. However, SCMPC typically ignores the structural relationships among scenarios and requires numerous samples to provide safety guarantees, resulting in a high computational cost.
Stochastic MPC (SMPC) \citep{carvalho2014stochastic, brudigam2021stochastic} replaces hard constraints with chance constraints that allow a small probability of constraint violation, thereby reducing trajectory-level conservatism, i.e., conservatism associated with trajectory uncertainty.
To handle intention uncertainty, the method in \cite{benciolini2023non} prioritizes safety constraints across future intentions based on  their prediction probabilities, avoiding excessive restrictions from those that are considered unlikely in multi-modal prediction. 
Nevertheless, SMPC typically does not anticipate how future information will reduce SVs' uncertainty, limiting its ability to plan less conservatively. At each planning iteration, SMPC generates a single control sequence that must remain feasible and safe under chance constraints across all possible intentions over the entire prediction horizon. This neglects the fact that each SV will ultimately commit to a specific behavior in the near future, meaning that intention uncertainty naturally decreases when new observations become available \citep{isele2025delayed}.

To anticipate future feedback from SVs, \cite{nair2024predictive} propose a feedback policy that enables responses to different realizations of SV behaviors over the prediction horizon, thus reducing conservatism.
Branch MPC (BMPC) \citep{chen2022interactive}, also known as contingency MPC \citep{alsterda2019contingency}, plans multiple contingency trajectories in parallel for different prediction scenarios using a branching structure.
By applying common control inputs across all branches until a designated branching time, BMPC allows the AV to postpone the commitment to a particular trajectory until additional information reduces intention uncertainty \citep{peters2024contingency}.

Despite these advantages, existing BMPC methods, which enumerate all possible joint combinations of intentions across SVs, suffer from combinatorial complexity. The number of combinations grows exponentially with the number of SVs, leading to high-dimensional, computationally challenging optimization problems.
Recently, scenario selection methods alleviate this by pruning the scenario tree, but risk discarding safety-critical cases \citep{fors2022resilient}. 
In \cite{bouzidi2025reachability}, the planning safety is guaranteed by extracting and merging driving corridors for each prediction scenario based on reachability analysis. 
However, merging scenarios based on corridor similarity does not necessarily preserve the optimality of the original problem, since scenarios with similar feasible corridors may still induce different optimal AV decisions.
Moreover, many BMPC formulations do not explicitly account for SVs' trajectory uncertainty, which  may compromise safety when actual SV behaviors deviate from nominal predictions.
Although some works incorporate trajectory uncertainty via branch weighting or risk-aware objectives \citep{chen2022interactive, zhang2024efficient}, these approaches only indirectly influence safety through the cost function and lack explicit probabilistic guarantees compared to SMPC.

\textit{Contribution:} In this work, we introduce a novel branch-stochastic MPC (B-SMPC) framework that enables safe, non-conservative, and computationally efficient motion planning for autonomous driving under multi-modal uncertainty.
By combining SMPC with the branching structure of BMPC, our approach can plan distinct trajectories for different SV intentions while ensuring safety with respect to trajectory uncertainty. 
We first construct a branching structure based on BMPC to handle intention uncertainty. 
Safety constraints are then formulated as chance constraints, enabling the optimization to explicitly incorporate trajectory uncertainty.
To address the combinatorial complexity of the branching structure in multi-vehicle scenarios, we propose a scenario-clustering method based on high-level maneuver planning for the AV. For each prediction scenario, a lightweight maneuver planner generates an optimal high-level maneuver, including target speed and target lateral position. Scenarios that lead to similar AV maneuvers are merged, forming clusters that represent unified maneuver options and significantly reducing the number of branches. 
Furthermore, since determining when to commit to a single trajectory affects the feasibility and performance of planning \citep{peters2024contingency}, we determine it adaptively using scenario similarity based on the Dynamic Time Warping (DTW) metric \citep{muller2007dynamic} to ensure safe and efficient decision postponement. 

\textit{Structure:} The remainder of the paper is organized as follows. Section~\ref{sec: problem formulation} formulates the motion planning problem under multi-modal uncertainty. Section~\ref{sec:preliminaries} presents the necessary preliminaries, including vehicle models, multi-modal trajectory prediction, and SMPC. In Section~\ref{sec: main part}, we introduce the proposed motion planning framework with scenario clustering and adaptive branching time computation. 
Section~\ref{sec: experiments} demonstrates the performance of the proposed framework through simulation experiments. Finally, Section~\ref{sec: conclusion} concludes the paper and discusses directions for future work.
\section{Problem Formulation}
\label{sec: problem formulation}

This work focuses on motion planning problems for AVs in a dynamic traffic environment populated by multiple SVs whose future behaviors are uncertain and potentially interactive. These interactive behaviors and uncertainties of SVs can be captured by a probabilistic trajectory prediction approach, such as \cite{salzmann2020trajectron++} and \cite{lefkopoulos2020interaction}. 

At each time step, the predictor observes the states of SVs and outputs a set of prediction modes for each SV $o \in \mathcal{O} = \{1, \dots, N_{\mathrm{SV}}\}$. 
Each prediction mode $i$ represents a plausible intention of the SV $o$, with all modes collected in the set $\Theta^{o}$.
For every prediction mode $i$ of the SV $o$, the predictor outputs the nominal predicted trajectory $\bm X_{}^{o, i}$ with associated mode probability $\mu_{}^{o, i}$ and predicted state covariance sequence $\bm P_{}^{o, i}$ over the prediction horizon $N$, representing intention and trajectory uncertainty, respectively. The complete prediction is denoted as
\begin{equation} \label{eq: prediction}
    \mathcal{P}^{o,i} = (\mu_{}^{o, i}, \bm X_{}^{o, i}, \bm P_{}^{o, i}),
\end{equation}
where $\bm X_{}^{o, i} = \{\bm x_{k}^{o, i}\}, \forall k \in [0, N]$ with the predicted state $\bm x_{k}^{o, i}$, and $\bm P_{}^{o, i} = \{\bm \Sigma_{k}^{o, i}\}, \forall k \in [0, N]$ with the covariance matrix $\bm{\Sigma}^{o,i}_{k}$ of the predicted state $\bm x_{k}^{o, i}$.
The multi-modal prediction set for the SV $o$ is defined as $\mathcal{P}^{o}= \{\mathcal{P}^{o,i}|i\in \Theta^o\}$.
Given these multi-modal predictions, the AV must plan a trajectory that ensures safety under multi-modal uncertainty while maintaining efficiency and comfort. The trajectory must also be real-time feasible within a receding-horizon framework, enabling the AV to apply the computed controls and continuously replan.

\section{Preliminaries}
\label{sec:preliminaries}

\subsection{Vehicle Models} \label{sec:vehicle model}
We employ two different motion models for motion planning and high-level maneuver planning to balance modeling accuracy and computational efficiency. 

\textit{Kinematic Bicycle Model:} The focus of the motion planning considered in this work lies in path and speed planning. For this purpose, a kinematic bicycle model \citep{rajamani2011vehicle} is adopted, which inherently satisfies steering angle constraints and provides sufficient accuracy under normal driving conditions with moderate accelerations. Let the state vector $\bm \xi = [x, y, \phi, v, a, \delta]^{\top}$, where $[x, y]^{\top}$ denotes the vehicle's position in the global frame, $\phi$ the yaw angle, $v$ the velocity, $a$ the acceleration, and $\delta$ the steering angle. The control input vector is $\bm u = [\dot{a}, \dot{\delta}]^{\top}$, comprising the jerk $\dot{a}$ and the steering rate $\dot{\delta}$. The continuous-time dynamics are described as
\begin{equation}\label{eq: continunous dynamics}
\dot{\boldsymbol{\xi}}
 = \bm f (\bm \xi, \bm u) = [\,v\cos\psi,\; v\sin\psi,\; v\tan\delta/L,\; a,\; \dot{a},\; \dot{\delta}\,]^\top,
\end{equation}
where $L$ is the wheelbase of the vehicle.
For numerical implementation in the MPC framework, we discretize (\ref{eq: continunous dynamics}) using Euler integration with a given time step size $dt$: 
\begin{equation} \label{eq: discretized dynamics}
    \bm \xi_{k+1} = \bm f_d(\bm \xi_{k}, \bm u_{k}) = \bm \xi_{k} + \bm f (\bm \xi_{k}, \bm u_{k}) dt.
\end{equation}

\textit{Simplified Vehicle Model:}
For maneuver planning, the motion model is represented as a simple point-mass model with decoupled longitudinal and lateral dynamics along the centerline of the target lane $\ell$. Let $\bm x^{\ell} = [s, \dot{s}, \ddot{s}, d, \dot{d}, \ddot{d}]^\top$ denote the state vector for the point-mass model, consisting of longitudinal and lateral positions, velocities, and accelerations, respectively.
The input $\bm u^{\ell} = [j_s, j_d]$ comprises the longitudinal and lateral jerk.
We model vehicle actions using an LQR-based controller that tracks reference targets defining a maneuver. In this work, a maneuver $\mathfrak{m}^{\ell}$ is specified by a longitudinal target speed $v^{\ell}_{\text{ref}}$ and a lateral target position $d^{\ell}_{\text{ref}}$ along the target lane $\ell$, and is denoted as 
\begin{equation} \label{eq: maneuver}
    \mathfrak{m}^{\ell} = (\ell, v^{\ell}_{\mathrm{ref}}, d^{\ell}_{\mathrm{ref}}).
\end{equation}
The motion model for the maneuver $\mathfrak{m}^{\ell}$ is given by
\begin{equation} \label{eq: pm model with lqr}
    \bm x_{k+1}^{\ell} = \bm{A} \bm x_{k}^{\ell} + \bm{B} \bm{u}_{k}^{\ell}, \quad \bm u_{k}^{\ell} = \bm K^{\ell}(\bm x_{k}^{\ell} - \bm x_{\mathrm{ref}}^{\ell}(\mathfrak{m}^{\ell})),
\end{equation}
where $\bm A$ and $\bm B$ are the discrete-time point-mass model matrices as in \cite[(2)]{zhou2023interaction}, and the control input $\bm u_{k}^{\ell}$ is determined by a feedback loop based on the state reference $\bm x_{\mathrm{ref}}^{\ell}$ generated from the maneuver $\mathfrak{m}^{\ell}$, and the control gain $\bm K^{\ell}$ obtained by LQR.

\subsection{Multi-modal Trajectory Prediction} \label{sec:iaimm-kf}
A probabilistic prediction approach is required to account for the inherent uncertainty in the intentions and trajectories of SVs. In this work, we use the interaction-aware interacting multiple model Kalman filter (IAIMM-KF) \citep{lefkopoulos2020interaction}, a model-based interaction-aware motion-prediction method. 
The IAIMM-KF provides multi-modal trajectory predictions that capture interactions and multi-modal uncertainties of SVs. Based on interactive motion predictions from the IAIMM-KF, our motion planning framework can achieve interaction awareness and perform more proactively in multi-vehicle scenarios, as verified in  \cite{zhou2023interaction}. 
Prediction modes in IAIMM-KF are defined as different dynamic models, such as (\ref{eq: pm model with lqr}), to represent different driving intentions of SVs. The probability of each prediction mode is recursively updated based on incoming observations and vehicle-to-vehicle interactions.
For the theoretical details and the complete algorithm, we refer interested readers to \cite{lefkopoulos2020interaction}.

Based on the IAIMM-KF outputs, we construct a set of prediction scenarios $\Omega$ by combining the prediction modes $\Theta^{o}$ of each SV $o \in \mathcal{O}$. 
The overall scenario set is defined as the Cartesian product $\Omega = \Theta^1 \times \cdots \times\Theta^{N_{\mathrm{SV}}}$.
Each scenario $\omega \in \Omega$ represents a distinct joint realization of all SV behaviors, with scenario probability $p_{\omega}$ given by the product of the individual mode probabilities. The predictions under scenario $\omega$ are denoted as $\mathcal{P}_{\omega}$, comprising the predictions $\mathcal{P}^{o,i}$ for all SVs in scenario $\omega$.

\subsection{Stochastic Model Predictive Control}
In the SMPC framework, the AV plans its motion by solving a finite-horizon optimal control problem (OCP). 
The optimal trajectory is chosen by minimizing a cost function while hard constraints for collision avoidance are relaxed and must be satisfied only up to a certain level of probability with a confidence parameter $0 < \beta <1$. 
Only the first element of the planned trajectory is applied to the system, and the process is repeated over a receding horizon.
The SMPC OCP is formulated as:
\begin{subequations}\label{eq: smpc}
    \begin{align}
        &\min_{\bm U} \quad  J(\bm \xi_0, \bm U)  \label{eq:smpc cost}\\
\text{s.t.} \quad 
& \bm \xi_{k+1} = f_d(\bm \xi_k, \bm u_k), \quad k \in [0, N-1], \label{eq: smpc dynamic}\\
& \bm \xi_k \in \mathcal{X}_k, \quad\quad\quad\quad\quad k \in [1, N], \label{eq: smpc state cons}\\
& \bm u_k \in \mathcal{U}_k, \quad\quad\quad\quad\quad k \in [0, N-1], \label{eq: smpc input cons}\\
& \Pr[\bm \xi_k \in \mathcal{S}^{o, i}] \ge \beta^{o, i}, k \in [1, N], i\in \Theta^o, o \in \mathcal{O}, \label{eq: smpc chance constraint}
    \end{align}
\end{subequations}
where $N$ is the prediction horizon, $\bm U = \{\bm u_k\}_{k=0}^{N-1}$ denotes the sequence of the control inputs, 
$\mathcal{X}_k$ and $\mathcal{U}_k$ are state and input hard constraints considering the physical limitations of the vehicle, 
and $\mathcal{S}^{o, i}$ is the safe state set w.r.t. the SV $o$ and its prediction mode $i \in \Theta^{o}$. 
The cost $J(\bm \xi_0, \bm U)$ is designed to penalize large inputs and large deviations $\Delta \bm \xi = \bm \xi - \bm \xi^{\ast}_{\mathfrak{m}^{\ell}}$ of the AV from the reference state $\bm \xi^{\ast}_{\mathfrak{m}^{\ell}}$ associated with maneuver $\mathfrak{m}^{\ell}$
\begin{equation} \label{eq:cost function definition}
    J(\bm \xi_0, \bm U) = \left \| \Delta \bm \xi_N \right \|_{\bm Q} + \sum_{k=0}^{N-1} (\left \| \Delta \bm \xi_k \right \|_{\bm R} + \left \| \bm u_k \right \|_{\bm S}),
\end{equation}
with weighting matrices $\bm Q, \bm R \succeq  0 , \bm S \succ 0$. 
Following our previous work, \cite{benciolini2023non}, the chance constraints (\ref{eq: smpc chance constraint}) guarantee safety with a confidence parameter $\beta^{o, i}$, which is adapted according to the corresponding mode probability $\mu_{}^{o, i}$ to reduce conservatism arising from trajectory uncertainty
\begin{equation} \label{eq: beta}
    \beta^{o, i} = g(\mu_{}^{o, i}),
\end{equation}
where $g(\cdot)$ is a monotonically increasing mapping, e.g., $g(p) = p^{\phi}$ with $0 < \phi \le 1$. 
This adaptation ensures that highly probable SV behaviors impose stronger safety guarantees, while less likely intentions are treated with reduced weight.

\section{Branch-Stochastic Model Predictive Control}
\label{sec: main part}

\begin{figure}[t!] 
    \centering 
    \footnotesize
\begingroup%
  \makeatletter%
  \providecommand\color[2][]{%
    \errmessage{(Inkscape) Color is used for the text in Inkscape, but the package 'color.sty' is not loaded}%
    \renewcommand\color[2][]{}%
  }%
  \providecommand\transparent[1]{%
    \errmessage{(Inkscape) Transparency is used (non-zero) for the text in Inkscape, but the package 'transparent.sty' is not loaded}%
    \renewcommand\transparent[1]{}%
  }%
  \providecommand\rotatebox[2]{#2}%
  \newcommand*\fsize{\dimexpr\f@size pt\relax}%
  \newcommand*\lineheight[1]{\fontsize{\fsize}{#1\fsize}\selectfont}%
  \ifx\svgwidth\undefined%
    \setlength{\unitlength}{243.36613452bp}%
    \ifx\svgscale\undefined%
      \relax%
    \else%
      \setlength{\unitlength}{\unitlength * \real{\svgscale}}%
    \fi%
  \else%
    \setlength{\unitlength}{\svgwidth}%
  \fi%
  \global\let\svgwidth\undefined%
  \global\let\svgscale\undefined%
  \makeatother%
  \begin{picture}(1,0.48437276)%
    \lineheight{1}%
    \setlength\tabcolsep{0pt}%
    \put(0,0){\includegraphics[width=\unitlength,page=1]{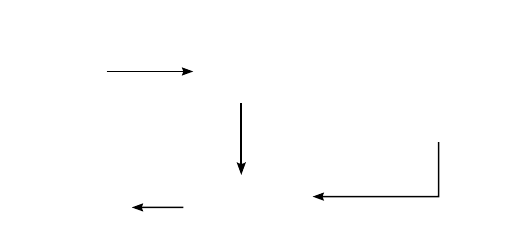}}%
    \put(0.88974565,0.29415383){\makebox(0,0)[t]{\lineheight{1.25}\smash{\begin{tabular}[t]{c}$k_{\mathrm{branch}}$\end{tabular}}}}%
    \put(0,0){\includegraphics[width=\unitlength,page=2]{framework.pdf}}%
  \end{picture}%
\endgroup%

    \caption{Overview of motion planning framework.} 
    \label{fig:overview} 
\end{figure}

In this section, we outline the proposed B-SMPC framework.
To handle both intention and trajectory uncertainty and overcome conservatism, we combine SMPC with BMPC and extend (\ref{eq: smpc}) by introducing a branching structure. 

The overview of our approach is shown in Fig.~\ref{fig:overview}. 
Based on the prediction scenarios $\Omega$ from the predictor, we construct a branching structure $\mathcal{B}$ that enables the planner to compute distinct trajectories for different SV behaviors. 
To ensure computational tractability, scenarios that induce similar AV maneuvers are clustered and merged into a single planning branch. Each branch $b \in \mathcal{B}$ therefore corresponds to a
cluster of scenarios $\mathcal{P}_b = \{\mathcal{P}_{\omega}\}$, together with its representative AV maneuver $\mathfrak{m}_b$ and its branch probability $\pi_b$. The scenario-clustering method will be detailed in Section~\ref{sec: Scenario Clustering}.

Incorporating non-anticipatory constraints into the OCP is another key feature of BMPC. Instead of making a decision immediately by choosing the most likely future, the planner first follows a trajectory that guarantees safety and efficiency for all candidate futures, and postpones the final decision until the uncertainty in SV behaviors has been sufficiently reduced.
The time at which the planner should commit to a specific branch is referred to as the branching time $k_\mathrm{branch}$. 
Choosing a branching time $k_\mathrm{branch}$ that is too small forces premature commitment and may compromise safety or efficiency, whereas choosing it too large reduces flexibility and may miss the opportunity to select the most appropriate maneuver. Our adaptive computation of the branching time will be detailed in Section~\ref{sec: branching time}.

By extending (\ref{eq: smpc}) through the branching structure $\mathcal{B}$ and non-anticipatory constraints up to the branching time $k_\mathrm{branch}$, the OCP in our framework is formulated as
\begin{subequations}\label{eq: bmpc}
    \begin{align}
        & \hspace{-0.5cm}\min_{\bm U_{b}, \forall b \in \mathcal{B}} \quad  \sum_{b\in \mathcal{B}} \pi_b J(\bm \xi_0, \bm U_{b}) \label{eq:bmpc cost}\\
\text{s.t.} \quad 
& \bm \xi_{k+1,b} = f_d(\bm \xi_{k,b}, \bm u_{k,b}), \quad k \in [0, N-1], b\in \mathcal{B}, \label{eq: bmpc dynamic}\\
& \bm \xi_{0, b} = \bm \xi_0, \quad b \in \mathcal{B}, \label{eq: bmpc initial cons}\\
& \bm \xi_{k,b} \in \mathcal{X}_k, \quad k \in [1, N], b \in \mathcal{B}, \label{eq: bmpc state cons}\\
& \bm u_{k,b} \in \mathcal{U}_k, \quad k \in [0, N-1], b \in \mathcal{B}, \label{eq: bmpc input cons}\\
& \Pr[\bm \xi_{k,b} \in \mathcal{S}^{o,i}] \ge \beta^{o,i}, 
  \notag \\
& \quad k \in [1,N], b \in \mathcal{B}, i \in \Theta_b^o, o \in \mathcal{O} \label{eq:bmpc chance} 
\\
& \bm u_{k, b_1} = \bm u_{k, b_2}, \quad k\in[0, k_{\mathrm{branch}}], (b_1, b_2) \in \mathcal{B}, \label{eq: bmpc branch cons}
    \end{align}
\end{subequations}
where $\bm U_b = \{\bm u_{k,b} \}_{k=0}^{N-1}$ is the control inputs of the AV for each branch. The objective of the OCP is to minimize the weighted sum of the costs across all branches, where each branch is weighted by its probability $\pi_b$. The states $\bm \xi_{k,b}$ should satisfy the chance constraints (\ref{eq:bmpc chance}) w.r.t. all prediction modes $\Theta_b^o$ for SV $o$ assigned to branch $b$, similar to (\ref{eq: smpc chance constraint}).
The non-anticipatory constraints (\ref{eq: bmpc branch cons}) enforce that all branches share a common control sequence before the branching time $k_\mathrm{branch}$, when the true situation is still uncertain. After this time, as the intention uncertainty of SVs has been sufficiently reduced through newly obtained observations, each branch is allowed to evolve independently. This enables the planner to adapt the subsequent trajectory to the anticipated behavior in each branch. Similar to \cite{bouzidi2025reachability}, we consider only a single branching point into different branches. 

\subsection{Scenario Clustering based on Maneuver Planning} \label{sec: Scenario Clustering}
In standard BMPC approaches, each branch corresponds to a single prediction scenario \citep{chen2022interactive, oliveira2023interaction}. 
However, since the number of scenarios grows exponentially with the number of SVs, the number of optimization variables in (\ref{eq: bmpc}) also scales exponentially, making standard BMPC computationally intractable for dense traffic environments. 
To address this problem, observing that many scenarios lead the AV to similar high-level maneuvers, we introduce a scenario-clustering method according to the AV's maneuver determined by the high-level maneuver planner, thereby reducing the number of branches without sacrificing decision quality.

\textit{Maneuver Planning}: For each prediction scenario $\omega \in \Omega$, the maneuver planner first generates the set of candidate collision-free maneuvers $\mathcal{M}_{\omega}$. A maneuver is considered collision-free if the trajectory yielded by tracking its target states satisfies all collision-avoidance conditions w.r.t. the nominal predicted trajectories of any SV in the scenario. Among these collision-free maneuvers, the maneuver planner selects the optimal maneuver by minimizing a maneuver cost function that reflects comfort and efficiency.

Following the maneuver definition in (\ref{eq: maneuver}), we identify a set of reachable lanes $\mathcal{L}$ from the AV's current state. Each reachable lane $\ell \in \mathcal{L}$ defines a corresponding maneuver $\mathfrak{m}^{\ell} \in \mathcal{M}$.
For each maneuver, the collision-free target longitudinal speed $v_{\mathrm{ref}}^{\ell}$ and target lateral position $d_{\mathrm{ref}}^{\ell}$ are obtained by solving a lightweight optimization problem
\begin{subequations}\label{eq:ego decision making}
\begin{align}
& \hspace{-0.5cm} \min_{v_{\mathrm{ref}}^{\ell}, d_{\mathrm{ref}}^{\ell}}\quad 
 W_v(v_{\mathrm{ref}}^{\ell} - \tilde{v}_{\mathrm{ref}}^{\ell})^2 + W_d{(d_{\mathrm{ref}}^{\ell})}^2 \label{eq:dm cost}\\
\text{s.t.}\quad
& \text{model in (\ref{eq: pm model with lqr})}, \bm x_{0}^{\ell}={\bm x}_{0}, \bm x_{k}\in \mathcal S\bigl(\mathcal{P}_{\omega}\bigr) \ \forall k\in[1, N], \label{eq:dm collision cons} 
\end{align}
\end{subequations}
where $W_v$ and $W_d$ are the weight parameters, $\tilde{v}_{\mathrm{ref}}^{\ell}$ is the speed limit of the target lane $\ell$.  
Safety constraints (\ref{eq:dm collision cons}) ensure collision avoidance with any SV following its nominal predicted trajectory in predictions $\mathcal{P}_{\omega}$ of scenario $\omega$. 
In practice, these safety constraints amount to enforcing non-overlap between the geometric occupancies of the AV and SVs.
\begin{rem}
    Problem (\ref{eq:ego decision making}) is computationally lightweight, as it involves only two optimization variables. Its real-time solvability has been demonstrated in \cite{lefkopoulos2020interaction}. In addition, as computations of collision-free maneuvers are independent, we parallelize their execution to further improve computational efficiency.
\end{rem}
After generating the set of candidate collision-free maneuvers $\mathcal{M}_{\omega}$, we evaluate the maneuver cost of each maneuver, accounting for comfort and efficiency, through \citep{zhou2022interaction}
\begin{equation} \label{eq: maneuver cost}
J_{\mathfrak{m}_{\omega}^{\ell}} 
= \sum_{k=1}^{N} 
\big(W_{\ddot{s}} (\ddot{s}_{k}^{\ell})^{2} + W_{\ddot{d}} (\ddot{d}_{k}^{\ell})^{2}\big) \\
+ J_{\Delta \mathrm{ref}} ,
\end{equation}
where $W_{\ddot{s}} , W_{\ddot{d}} , W_v$, and $W_d$ are weight parameters, and the cost $J_{\Delta \mathrm{ref}}$ is from (\ref{eq:dm cost}).
Then, we choose the maneuver with the lowest maneuver cost as the optimal maneuver $\mathfrak{m}_{\omega}^{\star}$
\begin{equation}
    \mathfrak{m}_{\omega}^{\star} = {\arg \min}_{\mathfrak{m}_{\omega}^{\ell} \in \mathcal{M}_{\omega}}(J_{\mathfrak{m}_{\omega}^{\ell}}).
\end{equation}
The remaining maneuvers are retained as backup options in case the scenario clustering or the OCP (\ref{eq: bmpc}) yields infeasible solutions.

\textit{Scenario Clustering}: A two-stage clustering method is employed to cluster prediction scenarios according to the similarity of the corresponding optimal AV maneuver $\mathfrak{m}_{\omega}^{\star}$. At the first stage, since the target lane determines the qualitative nature of the maneuver, e.g., lane keeping, lane change left or right, all scenarios are first partitioned into target-lane-specific subsets $\Omega_\ell = \{\, \omega \in \Omega \mid \ell_{\mathfrak{m}_\omega^\star} = \ell \,\},
\quad \ell \in \mathcal{L}.$
This stage ensures that only maneuvers within the same lane are compared in the subsequent clustering process. 
Then, maneuvers within each subset $\Omega_{\ell}$ are clustered in the standardized two-dimensional feature space of target speed and lateral position. 
The density-based spatial clustering of applications with noise (DBSCAN) algorithm \citep{ester1996density} is adopted to automatically identify compact groups of similar maneuvers, forming the cluster $\mathcal{C}_b$.
For each resulting cluster $\mathcal{C}_b$ with the corresponding target lane $\ell_b$, the target longitudinal velocity and target lateral position in the representative maneuver $\mathfrak{m}_{b}$ are defined as
\begin{equation}
\bar{v}^{\ell_b}_{\mathrm{ref}, b} = \frac{1}{|\mathcal{C}_b|} 
\sum_{\omega \in \mathcal{C}_b} v^{\ell_b}_{\mathrm{ref}, \omega}, 
\bar{d}^{\ell_b}_{\mathrm{ref}, b} = \frac{1}{|\mathcal{C}_b|} 
\sum_{\omega \in \mathcal{C}_b} d^{\ell_b}_{\mathrm{ref}, \omega}.
\label{eq:cluster_avg}
\end{equation}
The probability of the cluster is aggregated by prediction scenarios in this cluster, as $\pi_b = \sum_{\omega\in \mathcal{C}_b}p_{\omega}$.
Each resulting cluster corresponds to a branch $b$ in the branching structure $\mathcal{B}$ used in the OCP (\ref{eq: bmpc}). 

\begin{figure}[t!]
	\centering
	\def\svgwidth{1.0\columnwidth}
	\begin{subfigure}[t]{\columnwidth}
		\centering\scriptsize
		\def\svgwidth{\linewidth}
\begingroup%
  \makeatletter%
  \providecommand\color[2][]{%
    \errmessage{(Inkscape) Color is used for the text in Inkscape, but the package 'color.sty' is not loaded}%
    \renewcommand\color[2][]{}%
  }%
  \providecommand\transparent[1]{%
    \errmessage{(Inkscape) Transparency is used (non-zero) for the text in Inkscape, but the package 'transparent.sty' is not loaded}%
    \renewcommand\transparent[1]{}%
  }%
  \providecommand\rotatebox[2]{#2}%
  \newcommand*\fsize{\dimexpr\f@size pt\relax}%
  \newcommand*\lineheight[1]{\fontsize{\fsize}{#1\fsize}\selectfont}%
  \ifx\svgwidth\undefined%
    \setlength{\unitlength}{300.00006104bp}%
    \ifx\svgscale\undefined%
      \relax%
    \else%
      \setlength{\unitlength}{\unitlength * \real{\svgscale}}%
    \fi%
  \else%
    \setlength{\unitlength}{\svgwidth}%
  \fi%
  \global\let\svgwidth\undefined%
  \global\let\svgscale\undefined%
  \makeatother%
  \begin{picture}(1,0.22087535)%
    \lineheight{1}%
    \setlength\tabcolsep{0pt}%
    \put(0,0){\includegraphics[width=\unitlength,page=1]{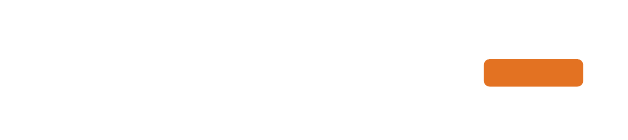}}%
    \put(0.85122901,0.09344098){\color[rgb]{0,0,0}\makebox(0,0)[t]{\lineheight{1.25}\smash{\begin{tabular}[t]{c}Cluster 2\end{tabular}}}}%
    \put(0,0){\includegraphics[width=\unitlength,page=2]{scenario_merging_two_clusters.pdf}}%
    \put(0.0959681,0.01082079){\makebox(0,0)[t]{\lineheight{1.25}\smash{\begin{tabular}[t]{c}AV\end{tabular}}}}%
    \put(0,0){\includegraphics[width=\unitlength,page=3]{scenario_merging_two_clusters.pdf}}%
    \put(0.40072456,0.00953286){\makebox(0,0)[t]{\lineheight{1.25}\smash{\begin{tabular}[t]{c}SV1\end{tabular}}}}%
    \put(0,0){\includegraphics[width=\unitlength,page=4]{scenario_merging_two_clusters.pdf}}%
    \put(0.85122901,0.17217177){\color[rgb]{0,0,0}\makebox(0,0)[t]{\lineheight{1.25}\smash{\begin{tabular}[t]{c}Cluster 1\end{tabular}}}}%
  \end{picture}%
\endgroup%

		\caption{Prediction scenarios are clustered into two clusters.}
		\label{fig:scenario_merging_two_clusters}
	\end{subfigure}
	\begin{subfigure}[t]{\columnwidth}
		\centering\scriptsize
		\def\svgwidth{\linewidth}
\begingroup%
  \makeatletter%
  \providecommand\color[2][]{%
    \errmessage{(Inkscape) Color is used for the text in Inkscape, but the package 'color.sty' is not loaded}%
    \renewcommand\color[2][]{}%
  }%
  \providecommand\transparent[1]{%
    \errmessage{(Inkscape) Transparency is used (non-zero) for the text in Inkscape, but the package 'transparent.sty' is not loaded}%
    \renewcommand\transparent[1]{}%
  }%
  \providecommand\rotatebox[2]{#2}%
  \newcommand*\fsize{\dimexpr\f@size pt\relax}%
  \newcommand*\lineheight[1]{\fontsize{\fsize}{#1\fsize}\selectfont}%
  \ifx\svgwidth\undefined%
    \setlength{\unitlength}{300bp}%
    \ifx\svgscale\undefined%
      \relax%
    \else%
      \setlength{\unitlength}{\unitlength * \real{\svgscale}}%
    \fi%
  \else%
    \setlength{\unitlength}{\svgwidth}%
  \fi%
  \global\let\svgwidth\undefined%
  \global\let\svgscale\undefined%
  \makeatother%
  \begin{picture}(1,0.2208754)%
    \lineheight{1}%
    \setlength\tabcolsep{0pt}%
    \put(0,0){\includegraphics[width=\unitlength,page=1]{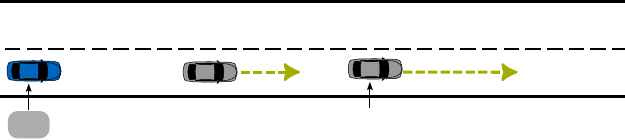}}%
    \put(0.04596819,0.0108208){\makebox(0,0)[t]{\lineheight{1.25}\smash{\begin{tabular}[t]{c}AV\end{tabular}}}}%
    \put(0,0){\includegraphics[width=\unitlength,page=2]{scenario_merging_one_cluster.pdf}}%
    \put(0.33072471,0.00953287){\makebox(0,0)[t]{\lineheight{1.25}\smash{\begin{tabular}[t]{c}SV1\end{tabular}}}}%
    \put(0,0){\includegraphics[width=\unitlength,page=3]{scenario_merging_one_cluster.pdf}}%
    \put(0.91624685,0.09381399){\color[rgb]{0,0,0}\makebox(0,0)[t]{\lineheight{1.25}\smash{\begin{tabular}[t]{c}Cluster 1\end{tabular}}}}%
    \put(0,0){\includegraphics[width=\unitlength,page=4]{scenario_merging_one_cluster.pdf}}%
    \put(0.59245193,0.01419412){\makebox(0,0)[t]{\lineheight{1.25}\smash{\begin{tabular}[t]{c}SV2\end{tabular}}}}%
    \put(0,0){\includegraphics[width=\unitlength,page=5]{scenario_merging_one_cluster.pdf}}%
  \end{picture}%
\endgroup%

		\caption{Prediction scenarios are clustered into a single cluster.}
		\label{fig:scenario_merging_one_cluster}
	\end{subfigure}
	\caption{Illustration of the merged clusters obtained from scenario clustering. 
    Only a subset of scenarios from the full set $\Omega$ is shown. In these examples, a single SV exhibits different behaviors while all other SVs remain identical.
    }
	\label{fig:scenario_merging}
\end{figure}

It is worth noting that not all SVs influence the AV's high-level decision, and scenario merging typically arises in two situations, as illustrated in Fig.~\ref{fig:scenario_merging}. 
In the first situation, an SV (SV1) has multiple possible future behaviors, but some of them lead the AV to select similar maneuvers. As shown in Fig.~\ref{fig:scenario_merging_two_clusters}, if SV1 is predicted to accelerate or to change to the left lane, both behaviors cause the AV to keep its current lane, whereas a deceleration of SV1 triggers the AV's lane-changing maneuver. Scenarios that yield similar AV maneuvers are therefore merged, while those leading to different maneuvers remain separate.
In the second situation, an SV (SV2) is far ahead such that none of its possible behaviors affects the AV's lane-keeping or lane-changing decision, as illustrated
in Fig.~\ref{fig:scenario_merging_one_cluster}. Here, the AV's maneuver is affected by another closer SV (SV1). Therefore, all scenarios arising from SV2's behaviors can be merged into one cluster.
These examples indicate that an SV affects the AV's decision only if its possible intentions are distributed across different scenario clusters. This motivates separating SVs into critical and non-critical. 
An SV is critical if there exists at least one cluster $b\in \mathcal{B}$ in which only a subset of its prediction modes is present, i.e., when $|\Theta_b^o| < |\Theta^o|$. Such an SV exhibits intention uncertainty that alters the branching structure and must therefore be handled with greater caution. 
We denote the sets of critical and non-critical SVs by $\mathcal{O}^{\mathrm{crit}}$ and $\mathcal{O}^{\mathrm{ncrit}}$, respectively.
Formally, the separation principle is defined as 
\begin{subequations}\label{eq:critical_noncritical}
\begin{align}
    \mathcal{O}^{\mathrm{crit}} &=
    \{\,o \in \mathcal{O} \mid \exists\, b \in \mathcal{B}: |\Theta_b^o| < |\Theta^o| \}, \\
    \mathcal{O}^{\mathrm{ncrit}} &=
    \{\,o \in \mathcal{O} \mid \forall\, b \in \mathcal{B}: |\Theta_b^o| = |\Theta^o| \}.
\end{align}
\end{subequations}

\subsection{Chance Constraint Formulation} \label{sec: chance cons construction}
After constructing the branching structure, we use the chance constraints in (\ref{eq:bmpc chance}) to ensure collision-free motion of the AV in each branch under the trajectory uncertainty of SVs. Because such probabilistic
constraints cannot be handled directly in numerical optimization, we reformulate
them into tractable deterministic approximations. To balance safety and computational efficiency, we use different reformulations for critical and non-critical SVs.

For each critical SV $o \in \mathcal{O}^{\mathrm{crit}}$, prediction mode $i \in \Theta^o$, and time step $k\in[1,N]$, we impose a chance constraint using an elliptical approximation following \cite{benciolini2023non}.
The chance constraint in (\ref{eq:bmpc chance}) is formulated as \cite[(20)]{benciolini2023non}
\begin{equation}\label{eq:ellipse constraint}
q_k^{o,i}(\boldsymbol{\xi}_k) \le 0,
\end{equation}
which enforces that the AV state $\bm \xi_k$ lies outside the confidence ellipse $\mathcal{E}_k^{o,i}$. The ellipse $\mathcal{E}_k^{o,i}$ is determined by the nominal prediction and covariance matrix in (\ref{eq: prediction}) and scaled by the confidence parameter $\beta^{o,i}$, which is computed by (\ref{eq: beta}).
This constraint guarantees that the planned trajectory remains collision-free w.r.t. all possible positions of the critical SV $o$, with a probability at least the confidence parameter $\beta^{o,i}$.

For non-critical SVs, we adopt a more efficient geometric approximation following \cite[Section V.B]{zhou2023interaction}. The prediction of each non-critical SV is represented as a Gaussian Mixture Model (GMM). The spatial occupancy region is extracted from the GMM using a tunable safety parameter, and over-approximated by a rectangular region. The corresponding chance constraint can be simplified to a single linear boundary of this rectangular occupancy region, which is selected based on the maneuver planning result in Section~\ref{sec: Scenario Clustering}. This yields convex, linear constraints that can be handled efficiently.

\subsection{Adaptive Branching Time from DTW Distance} \label{sec: branching time}

The branching time $k_{\mathrm{branch}}$ specifies when the AV may safely commit to a single planned trajectory.
Since this commitment is only safe once the intention uncertainty of SVs is sufficiently reduced, the branching time varies across traffic scenarios and must be estimated adaptively.
In this work, we compute it using a DTW-based similarity metric that quantifies the divergence among different SVs' intentions.

The DTW method is a widely used method for analyzing the similarity of time series data \citep{muller2007dynamic}. 
Unlike approaches that evaluate the overlap of predicted trajectories only in spatial space, DTW also accounts for temporal misalignment between trajectories through its alignment cost, i.e., the DTW distance, yielding a more robust metric of motion similarity. As noted in Sec. \ref{sec: Scenario Clustering}, since the non-critical SVs have negligible influence on the AV's decision, we compute the DTW distance only for critical SVs to improve the computational efficiency.
For each critical SV $o \in \mathcal{O}^{\mathrm{crit}}$, we consider a prediction mode pair $(m, n)$, where $m \in \Theta_b^o$, $n \in \Theta_{b'}^o$ are prediction modes in distinct branches $b, b' \in \mathcal{B}$, respectively. The local cost $c^{o, (m,n)}_{i, j}$ for the DTW metric between mode $m$ at time step $i$ and mode $n$ at time step $j$ is computed using the Mahalanobis distance:
\begin{equation}
    c^{o, (m,n)}_{i, j} = \sqrt{(\bm x_i^{o, m} - \bm x_j^{o, n})^\top {(\bm \Sigma^{(i,j)})}^{-1}(\bm x_i^{o, m} - \bm x_j^{o, n})},
\end{equation}
where $\bm \Sigma^{(i,j)} = ({\bm \Sigma_i^{o,m} + \bm \Sigma_j^{o,n}})/{2}$ with covariances $\bm \Sigma_i^{o,m}$ and $\bm \Sigma_j^{o,n}$, $\bm x_i^{o, m}$ and $\bm x_j^{o, n}$ are the nominal predicted states of modes $m$ and $n$. For each prediction time step $k \in [1, N]$, the DTW distance $\mathcal{D}^{o, (m,n)}_{k, k}$ is obtained by dynamic programming
\begin{equation}
    \mathcal{D}^{o, (m,n)}_{k, k} = c^{o, (m,n)}_{k, k} + \min(\mathcal{D}^{o, (m,n)}_{k-1, k}, \mathcal{D}^{o, (m,n)}_{k, k-1}, \mathcal{D}^{o, (m,n)}_{k-1, k-1}),
\end{equation}
The pairwise branching time for the prediction mode pair $(m, n)$ is then determined by
\begin{equation}
    k_{\mathrm{branch}}^{o, (m,n)} = \min \{k | \mathcal{D}^{o, (m,n)}_{k, k} \ge \Delta_{\mathrm{DTW}}\},
\end{equation}
where $\Delta_{\mathrm{DTW}}$ is a tunable threshold that specifies how much divergence between prediction modes is required before the planner commits to a single trajectory. A larger threshold corresponds to a more conservative assessment of uncertainty resolution, meaning that the prediction modes must have larger differences at the branching time. As a result, increasing $\Delta_{\mathrm{DTW}}$ delays the branching time within the same traffic scenario. 
The branching time $k_{\mathrm{branch}}^{o}$ for the critical SV $o$ is chosen as the maximum pairwise branching time over all prediction mode pairs drawn from distinct branches, ensuring that the planner remains non-anticipatory until the final possible divergence
\begin{equation}
    k_{\mathrm{branch}}^{o}
    = \max_{\substack{m \in \Theta_{b}^o,\; n \in \Theta_{b'}^o, b \neq b'}}
      k_{\mathrm{branch}}^{o,(m,n)}.
\end{equation}
If multiple critical SVs are present, the overall branching time is taken as $k_{\mathrm{branch}} = \max_{o\in \mathcal{O}^{\mathrm{crit}}}k_{\mathrm{branch}}^{o}$.

\begin{rem}
Standard DTW has quadratic complexity w.r.t. the trajectory length. To improve computational efficiency, the Most Significant Operation First (MSOF)-DTW method \citep{yulianto2019msof} can be used to compute lower and upper bounds on the DTW
distance and terminate early once it is clear whether the threshold
$\Delta_{\mathrm{DTW}}$ is exceeded, avoiding evaluation of the full cost
matrix.
\end{rem}

\section{Experiments}
\label{sec: experiments}

This section evaluates the proposed B-SMPC through highway simulations from the CommonRoad benchmark suite \citep{althoff2017commonroad}. 
Two baseline methods are used for comparison. The first is a Nominal MPC (NMPC), which considers only the most likely prediction of each SV in planning. The second baseline is an SMPC proposed by \cite{benciolini2023non}, which considers multi-modal uncertainty of SVs over the entire planning horizon and incorporates chance constraints for collision avoidance. Note that when the branching time $k_\mathrm{branch}$ is set identically to the full planning horizon $N$, SMPC becomes equivalent to B-SMPC. 

All planners are implemented in Python on a computer with an AMD Ryzen 7840HS CPU and 16 GB of memory. The associated nonlinear optimization problems are solved using CasADi \citep{andersson2019casadi} with the IPOPT solver \citep{wachter2006implementation}.
Both the prediction and planning time horizons are set to 30 time steps ($N = 30$), with a time step size $dt = 0.1\,\mathrm{s}$.
To ensure comparability, all planners use identical vehicle dynamics, cost weights, and actuation constraints.

\begin{figure}[t!]
    \centering
    \def\svgwidth{1.0\columnwidth}\footnotesize
\begingroup%
  \makeatletter%
  \providecommand\color[2][]{%
    \errmessage{(Inkscape) Color is used for the text in Inkscape, but the package 'color.sty' is not loaded}%
    \renewcommand\color[2][]{}%
  }%
  \providecommand\transparent[1]{%
    \errmessage{(Inkscape) Transparency is used (non-zero) for the text in Inkscape, but the package 'transparent.sty' is not loaded}%
    \renewcommand\transparent[1]{}%
  }%
  \providecommand\rotatebox[2]{#2}%
  \newcommand*\fsize{\dimexpr\f@size pt\relax}%
  \newcommand*\lineheight[1]{\fontsize{\fsize}{#1\fsize}\selectfont}%
  \ifx\svgwidth\undefined%
    \setlength{\unitlength}{300bp}%
    \ifx\svgscale\undefined%
      \relax%
    \else%
      \setlength{\unitlength}{\unitlength * \real{\svgscale}}%
    \fi%
  \else%
    \setlength{\unitlength}{\svgwidth}%
  \fi%
  \global\let\svgwidth\undefined%
  \global\let\svgscale\undefined%
  \makeatother%
  \begin{picture}(1,0.23939771)%
    \lineheight{1}%
    \setlength\tabcolsep{0pt}%
    \put(0,0){\includegraphics[width=\unitlength,page=1]{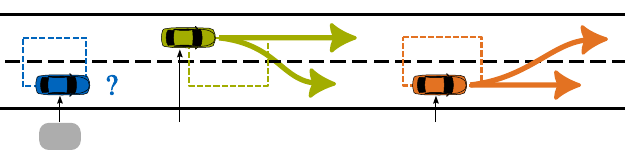}}%
    \put(0.09596814,0.00897055){\makebox(0,0)[t]{\lineheight{1.25}\smash{\begin{tabular}[t]{c}AV\end{tabular}}}}%
    \put(0,0){\includegraphics[width=\unitlength,page=2]{experiment_setting.pdf}}%
    \put(0.28818732,0.00953289){\makebox(0,0)[t]{\lineheight{1.25}\smash{\begin{tabular}[t]{c}SV1\end{tabular}}}}%
    \put(0,0){\includegraphics[width=\unitlength,page=3]{experiment_setting.pdf}}%
    \put(0.69811809,0.00953289){\makebox(0,0)[t]{\lineheight{1.25}\smash{\begin{tabular}[t]{c}SV2\end{tabular}}}}%
  \end{picture}%
\endgroup%

    \caption{Sketch of the traffic configuration used for simulations.}
    \label{fig:test_scenario}
\end{figure}

The evaluation scenario is a two-lane highway overtaking scenario in which the AV aims to safely and efficiently overtake the slower SVs ahead, as illustrated in Fig.~\ref{fig:test_scenario}. Each SV may keep its current lane or execute a lane-change maneuver, resulting in multi-modal uncertainty.
As described in Section~\ref{sec:iaimm-kf}, we use the IAIMM-KF as the trajectory predictor. Since the AV has no prior knowledge of the SVs' true intentions, the predictor initializes the intention probabilities uniformly at the beginning of each simulation.

The planners are evaluated through a Monte Carlo study.
Each simulation runs for 50 time steps ($5\,\mathrm{s}$) and operates in a receding-horizon fashion: at every time step, the planner receives updated multi-modal trajectory predictions from the predictor, solves a new OCP, and applies the first control input.
We repeat the simulations for 12 initial states of the AV and 30 initial states of SVs, generated from a uniform grid over the regions shown in Fig.~\ref{fig:test_scenario}.
The ground-truth trajectories of SVs are produced using the model in (\ref{eq: pm model with lqr}) with randomly sampled parameters. We first present two challenging scenarios to illustrate the behavioral differences between NMPC, SMPC, and B-SMPC. Then, we show quantitative results from Monte Carlo simulations.

\subsection{Qualitative Comparison}
\begin{figure}[t!]
	\centering
	\def\svgwidth{1.0\columnwidth}
	\begin{subfigure}[t]{\columnwidth}
		\centering\scriptsize
		\def\svgwidth{\linewidth}
		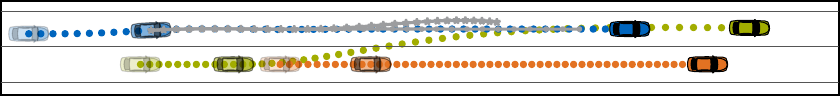
		\caption{B-SMPC.}
		\label{fig:result1_bmpc}
	\end{subfigure}
	\begin{subfigure}[t]{\columnwidth}
		\centering\scriptsize
		\def\svgwidth{\linewidth}
		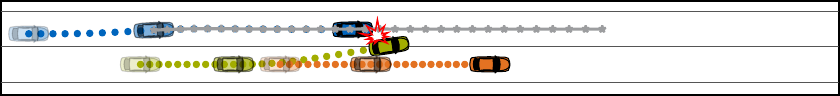
		\caption{NMPC.}
		\label{fig:result1_mpc}
	\end{subfigure}
	\caption{Closed-loop results for the first scenario, comparing B-SMPC and NMPC.}
	\label{fig:result1}
\end{figure}

In the first scenario, the AV is initialized in the left lane behind two slower vehicles, SV1 and SV2, in the right lane, whose intentions are initially ambiguous. SV2 stays in the right lane throughout, while SV1 initially remains in the right lane and begins changing to the left lane at the time $t = 1\,\mathrm{s}$. Fig.~\ref{fig:result1} shows the resulting closed-loop trajectories controlled by NMPC and B-SMPC, ground truth trajectories of SVs, and the planned trajectory of NMPC and B-SMPC at $t = 1\,\mathrm{s}$. NMPC initially accelerates optimistically to perform an early overtake, but SV1's unexpected lane change creates discrepancies between the predicted and realized SV behavior. From $t = 1.0\,\mathrm{s}$ to $t = 1.6\,\mathrm{s}$, the predictor continues to classify lane keeping as the most likely behavior of SV1. As a result, NMPC fails to anticipate SV1's lane change and maintains a high-speed plan, leading to a collision with SV1. Conversely, our B-SMPC maintains both the overtaking and yielding branches and postpones commitment until the prediction modes become sufficiently separated. Therefore, when SV1 changes to the left, B-SMPC safely commits to the yielding maneuver, allowing the AV to avoid excessive braking and prevent a collision.

\begin{figure}[t!]
	\centering
	\def\svgwidth{1.0\columnwidth}
	\begin{subfigure}[t]{\columnwidth}
		\centering\scriptsize
		\def\svgwidth{\linewidth}
		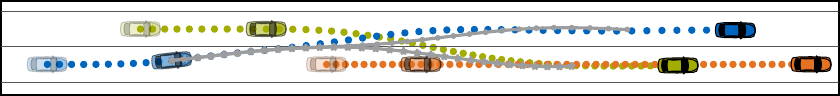
		\caption{B-SMPC.}
		\label{fig:result2_bmpc}
	\end{subfigure}
	\begin{subfigure}[t]{\columnwidth}
		\centering\scriptsize
		\def\svgwidth{\linewidth}
		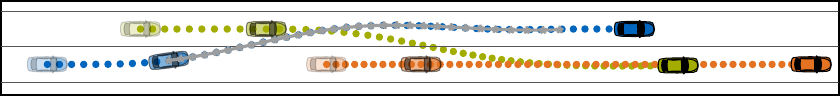
		\caption{SMPC.}
		\label{fig:result2_smpc}
	\end{subfigure}
	\caption{Closed-loop behavior in the second scenario, comparing B-SMPC and SMPC.}
	\label{fig:result2}
\end{figure}

In the second scenario, the AV is initialized in the right lane. Two slower vehicles, SV1 and SV2, are positioned ahead of the AV on the left and right lanes, respectively. SV2 remains in the right lane throughout, while SV1 initially stays in the left lane, then decelerates at $t = 1\,\mathrm{s}$ and begins changing to the right lane. Fig.~\ref{fig:result2} illustrates the resulting closed-loop trajectories controlled by SMPC and B-SMPC, ground truth trajectories of SVs, and the planned trajectory of SMPC and B-SMPC at $t = 1\,\mathrm{s}$. SMPC incorporates multi-modal uncertainty but must plan a single trajectory that remains feasible for all possible SV behaviors across the entire prediction horizon. To satisfy safety constraints, SMPC adopts an overly conservative strategy, accelerating only until the uncertainty fully resolves. Consequently, the AV follows the slower SVs cautiously for a long time before overtaking.
In contrast, our B-SMPC mitigates such conservatism by anticipating future SVs' feedback and allowing scenario-dependent behavior after the branching time. This flexibility enables B-SMPC to start the overtaking maneuver significantly earlier than SMPC, resulting in faster progress and smoother motion while maintaining safety. 

\subsection{Quantitative Analysis}
In addition to the two baseline planners, we include three ablation variants of B-SMPC to assess the contribution of each component. 
B-SMPC\textsubscript{noSC} removes the scenario clustering, so the number of branches equals the number of prediction scenarios.
BMPC considers only nominal trajectory predictions, thereby ignoring trajectory uncertainty.
B-SMPC\textsubscript{noDTW} disables the adaptive branching-time computation and instead fixes the branching time at $k_{\mathrm{branch}} = 2$. For all methods, we evaluate performance along three dimensions.
First, we measure planning-failure rates, including collisions and optimization infeasibilities that trigger a fail-safe planner \citep{brudigam2021stochastic}. 
Second, we assess comfort and efficiency using the cost terms defined in (\ref{eq:cost function definition}). 
Third, we evaluate the average computation time per planning step, accounting for the OCP solution time and any method-specific components, such as scenario clustering or adaptive branching-time computation, when applicable.

As summarized in table~\ref{tab:quantitative}, the proposed B-SMPC achieves the best overall balance among safety, efficiency, and computation time, outperforming all baseline methods. It obtains a zero failure rate, matching SMPC, while achieving a substantially lower cost. NMPC exhibits the highest failure rate because it relies solely on the most likely prediction. In contrast, although SMPC guarantees safety, its overly conservative strategy leads to the highest cost among all methods.
The ablation B-SMPC\textsubscript{noSC} highlights the importance of scenario clustering. Without scenario clustering, the planner must treat every prediction scenario as a separate branch, resulting in the largest computation time among the methods.
BMPC achieves the lowest cost among all baselines but suffers a higher failure rate because it ignores trajectory uncertainty. 
Most of its failures occur when SV motions deviate from the nominal predictions.
B-SMPC\textsubscript{noDTW} commits to a distinct trajectory too early due to its fixed branching time, leading to unsafe behavior and increased cost.

\begin{figure}[t!] 
    \centering 
    \def\svgwidth{0.7\columnwidth}\footnotesize
    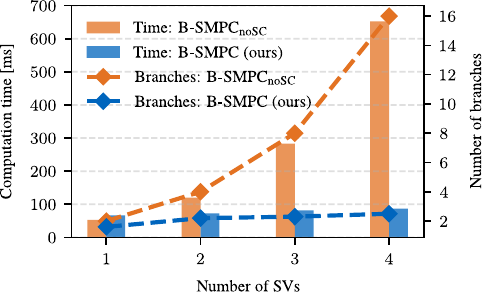
    \caption{Computation time and average branch count of our method and B-SMPC\textsubscript{noSC} with increasing number of SVs.} 
    \label{fig:scalability} 
\end{figure}

We further evaluate the computational scalability of our method by varying the number of SVs and comparing it with B-SMPC\textsubscript{noSC}. 
As shown in Fig.~\ref{fig:scalability}, the computation time of B-SMPC\textsubscript{noSC} increases sharply with the number of SVs, exceeding $600\,\mathrm{ms}$ for four SVs due to the rapid growth in the number of branches, which is up to $16$. In contrast, our B-SMPC keeps the number of branches small, yielding computation time below $100\,\mathrm{ms}$ even with four SVs.
By preventing the combinatorial growth in branches, scenario clustering enables B-SMPC to maintain real-time performance in multi-vehicle scenarios.

\begin{table}[t]
    \centering
    \footnotesize
    \caption{Results of Monte Carlo simulations.}
    \label{tab:quantitative}
    \begin{tabular}{lccc}
        \toprule
        \textbf{Method} & \textbf{Failure rate [\%]} & \textbf{Cost $\downarrow$} & \textbf{Time [ms]$\downarrow$} \\
        \midrule
        NMPC & 14.9 & 290.7 & \textbf{38} \\
        SMPC & 0.0 & 727.6 & 44 \\
        B-SMPC\textsubscript{noSC} & 0.0 & 119.4 & 116+4\\
        BMPC & 3.3 & \textbf{108.6} & 62+24 \\
        B-SMPC\textsubscript{noDTW} & 3.1 & 294.4 & 55+20 \\
        \textbf{B-SMPC} (ours) & \textbf{0.0} & 118.6 & 46+24 \\
        \bottomrule
    \end{tabular}
\end{table}

\section{Conclusion}
\label{sec: conclusion}

This paper proposes a motion-planning framework, B-SMPC, that combines SMPC and BMPC in a novel way to enable safe, non-conservative autonomous driving under multi-modal prediction uncertainty. By clustering prediction scenarios based on a high-level maneuver planner, the proposed B-SMPC efficiently reduces the number of branches, ensuring tractability even in multi-vehicle traffic scenarios. Safety is further enhanced through chance constraints that explicitly incorporate the trajectory uncertainty of SVs. Moreover, an adaptive branching-time computation ensures that branching occurs only when the intention uncertainty of SVs is sufficiently reduced, preventing both premature and overly delayed decisions.
Simulation results in multi-vehicle highway scenarios demonstrate that the proposed B-SMPC achieves a better balance among robustness, performance, and computation time, outperforming SMPC and BMPC applied in isolation.
Future work will focus on integrating explicit interaction modeling between the AV and SVs into the branching structure, learning the scenario-clustering and branching-time selection mechanisms from data, and extending the current framework to more complex urban environments.

\bibliography{references}

@article{benciolini2023non,
  title={Non-conservative trajectory planning for automated vehicles by estimating intentions of dynamic obstacles},
  author={Benciolini, Tommaso and Wollherr, Dirk and Leibold, Marion},
  journal={IEEE Transactions on Intelligent Vehicles},
  volume={8},
  number={3},
  pages={2463--2481},
  year={2023},
  publisher={IEEE}
}

@article{zhou2023interaction,
  title={Interaction-aware motion planning for autonomous vehicles with multi-modal obstacle uncertainty predictions},
  author={Zhou, Jian and Olofsson, Bj{\"o}rn and Frisk, Erik},
  journal={IEEE Transactions on Intelligent Vehicles},
  volume={9},
  number={1},
  pages={1305--1319},
  year={2023},
  publisher={IEEE}
}

@article{lefkopoulos2020interaction,
  title={Interaction-aware motion prediction for autonomous driving: A multiple model kalman filtering scheme},
  author={Lefkopoulos, Vasileios and Menner, Marcel and Domahidi, Alexander and Zeilinger, Melanie N},
  journal={IEEE Robotics and Automation Letters},
  volume={6},
  number={1},
  pages={80--87},
  year={2020},
  publisher={IEEE}
}

@inproceedings{carvalho2014stochastic,
  title={Stochastic predictive control of autonomous vehicles in uncertain environments},
  author={Carvalho, Ashwin and Gao, Yiqi and Lefevre, St{\'e}phanie and Borrelli, Francesco},
  booktitle={12th international symposium on advanced vehicle control},
  volume={9},
  year={2014}
}

@article{brudigam2021stochastic,
  title={Stochastic model predictive control with a safety guarantee for automated driving},
  author={Br{\"u}digam, Tim and Olbrich, Michael and Wollherr, Dirk and Leibold, Marion},
  journal={IEEE Transactions on Intelligent Vehicles},
  volume={8},
  number={1},
  pages={22--36},
  year={2021},
  publisher={IEEE}
}

@inproceedings{zhou2022interaction,
  title={Interaction-aware moving target model predictive control for autonomous vehicles motion planning},
  author={Zhou, Jian and Olofsson, Bj{\"o}rn and Frisk, Erik},
  booktitle={2022 European Control Conference (ECC)},
  pages={154--161},
  year={2022},
  organization={IEEE}
}

@article{wang2022social,
  title={Social interactions for autonomous driving: A review and perspectives},
  author={Wang, Wenshuo and Wang, Letian and Zhang, Chengyuan and Liu, Changliu and Sun, Lijun and others},
  journal={Foundations and Trends{\textregistered} in Robotics},
  volume={10},
  number={3-4},
  pages={198--376},
  year={2022},
  publisher={Now Publishers, Inc.}
}

@article{shi2022motion,
  title={Motion transformer with global intention localization and local movement refinement},
  author={Shi, Shaoshuai and Jiang, Li and Dai, Dengxin and Schiele, Bernt},
  journal={Advances in Neural Information Processing Systems},
  volume={35},
  pages={6531--6543},
  year={2022}
}

@INPROCEEDINGS{bouzidi2025reachability,
  author={Bouzidi, Mohamed-Khalil and Derajic, Bojan and Goehring, Daniel and Reichardt, Joerg},
  booktitle={2025 IEEE Intelligent Vehicles Symposium (IV)}, 
  title={Reachability-Based Contingency Planning Against Multi-Modal Predictions with Branch MPC}, 
  year={2025},
  volume={},
  number={},
  pages={1086-1093},
  doi={10.1109/IV64158.2025.11097372}}

@article{peters2024contingency,
  title={Contingency games for multi-agent interaction},
  author={Peters, Lasse and Bajcsy, Andrea and Chiu, Chih-Yuan and Fridovich-Keil, David and Laine, Forrest and Ferranti, Laura and Alonso-Mora, Javier},
  journal={IEEE Robotics and Automation Letters},
  volume={9},
  number={3},
  pages={2208--2215},
  year={2024},
  publisher={IEEE}
}

@inproceedings{levinson2011towards,
  title={Towards fully autonomous driving: Systems and algorithms},
  author={Levinson, Jesse and Askeland, Jake and Becker, Jan and Dolson, Jennifer and Held, David and Kammel, Soeren and Kolter, J Zico and Langer, Dirk and Pink, Oliver and Pratt, Vaughan and others},
  booktitle={2011 IEEE intelligent vehicles symposium (IV)},
  pages={163--168},
  year={2011},
  organization={IEEE}
}

@inproceedings{salzmann2020trajectron++,
  title={Trajectron++: Dynamically-feasible trajectory forecasting with heterogeneous data},
  author={Salzmann, Tim and Ivanovic, Boris and Chakravarty, Punarjay and Pavone, Marco},
  booktitle={European Conference on Computer Vision},
  pages={683--700},
  year={2020},
  organization={Springer}
}

@article{massera2017safe,
  title={Safe optimization of highway traffic with robust model predictive control-based cooperative adaptive cruise control},
  author={Massera Filho, Carlos and Terra, Marco H and Wolf, Denis F},
  journal={IEEE Transactions on Intelligent Transportation Systems},
  volume={18},
  number={11},
  pages={3193--3203},
  year={2017},
  publisher={IEEE}
}

@inproceedings{kensbock2023scenario,
  title={Scenario-based decision-making, planning and control for interaction-aware autonomous driving on highways},
  author={Kensbock, Robin and Nezami, Maryam and Schildbach, Georg},
  booktitle={2023 IEEE Intelligent Vehicles Symposium (IV)},
  pages={1--6},
  year={2023},
  organization={IEEE}
}

@article{cesari2017scenario,
  title={Scenario model predictive control for lane change assistance and autonomous driving on highways},
  author={Cesari, Gianluca and Schildbach, Georg and Carvalho, Ashwin and Borrelli, Francesco},
  journal={IEEE Intelligent transportation systems magazine},
  volume={9},
  number={3},
  pages={23--35},
  year={2017},
  publisher={IEEE}
}

@INPROCEEDINGS{isele2025delayed,
  author={Isele, David and Añon, Alexandre Miranda and Tariq, Faizan M. and Yeh, Goro and Singh, Avinash and Bae, Sangjae},
  booktitle={2025 IEEE International Conference on Robotics and Automation (ICRA)}, 
  title={Delayed-Decision Motion Planning in the Presence of Multiple Predictions}, 
  year={2025},
  volume={},
  number={},
  pages={11743-11749},
  doi={10.1109/ICRA55743.2025.11128178}}

@article{chen2022interactive,
  title={Interactive multi-modal motion planning with branch model predictive control},
  author={Chen, Yuxiao and Rosolia, Ugo and Ubellacker, Wyatt and Csomay-Shanklin, Noel and Ames, Aaron D},
  journal={IEEE Robotics and Automation Letters},
  volume={7},
  number={2},
  pages={5365--5372},
  year={2022},
  publisher={IEEE}
}

@inproceedings{alsterda2019contingency,
  title={Contingency model predictive control for automated vehicles},
  author={Alsterda, John P and Brown, Matthew and Gerdes, J Christian},
  booktitle={2019 American Control Conference (ACC)},
  pages={717--722},
  year={2019},
  organization={IEEE}
}

@article{fors2022resilient,
  title={Resilient branching MPC for multi-vehicle traffic scenarios using adversarial disturbance sequences},
  author={Fors, Victor and Olofsson, Bj{\"o}rn and Frisk, Erik},
  journal={IEEE Transactions on Intelligent Vehicles},
  volume={7},
  number={4},
  pages={838--848},
  year={2022},
  publisher={IEEE}
}

@inproceedings{zhang2024efficient,
  title={An Efficient Risk-aware Branch MPC for Automated Driving that is Robust to Uncertain Vehicle Behaviors},
  author={Zhang, Luyao and Pantazis, Georgios and Han, Shaohang and Grammatico, Sergio},
  booktitle={2024 IEEE 63rd Conference on Decision and Control (CDC)},
  pages={8207--8212},
  year={2024},
  organization={IEEE}
}

@book{rajamani2011vehicle,
  author    = {Rajamani, Rajesh},
  title     = {Vehicle Dynamics and Control},
  year      = {2011},
  publisher = {Springer},
  address   = {Berlin, Germany},
}

@inproceedings{ester1996density,
  title={A density-based algorithm for discovering clusters in large spatial databases with noise},
  author={Ester, Martin and Kriegel, Hans-Peter and Sander, J{\"o}rg and Xu, Xiaowei and others},
  booktitle={kdd},
  volume={96},
  number={34},
  pages={226--231},
  year={1996}
}

@inproceedings{yulianto2019msof,
  title={The MSOF-DTW Method for Checking Timeseries Similarities},
  author={Yulianto, Fazmah Arif and Mengko, Richard Karel Willem and others},
  booktitle={2019 7th International Conference on Information and Communication Technology (ICoICT)},
  pages={1--7},
  year={2019},
  organization={IEEE}
}

@inproceedings{althoff2017commonroad,
  title={CommonRoad: Composable benchmarks for motion planning on roads},
  author={Althoff, Matthias and Koschi, Markus and Manzinger, Stefanie},
  booktitle={2017 IEEE Intelligent Vehicles Symposium (IV)},
  pages={719--726},
  year={2017},
  organization={IEEE}
}

@article{andersson2019casadi,
  title={CasADi: a software framework for nonlinear optimization and optimal control},
  author={Andersson, Joel AE and Gillis, Joris and Horn, Greg and Rawlings, James B and Diehl, Moritz},
  journal={Mathematical Programming Computation},
  volume={11},
  number={1},
  pages={1--36},
  year={2019},
  publisher={Springer}
}

@article{wachter2006implementation,
  title={On the implementation of an interior-point filter line-search algorithm for large-scale nonlinear programming},
  author={W{\"a}chter, Andreas and Biegler, Lorenz T},
  journal={Mathematical programming},
  volume={106},
  number={1},
  pages={25--57},
  year={2006},
  publisher={Springer}
}

@article{liu2022interaction,
  title={Interaction-aware trajectory prediction and planning for autonomous vehicles in forced merge scenarios},
  author={Liu, Kaiwen and Li, Nan and Tseng, H Eric and Kolmanovsky, Ilya and Girard, Anouck},
  journal={IEEE Transactions on Intelligent Transportation Systems},
  volume={24},
  number={1},
  pages={474--488},
  year={2022},
  publisher={IEEE}
}

@inproceedings{oliveira2023interaction,
  title={Interaction and decision making-aware motion planning using branch model predictive control},
  author={Oliveira, Rui and Nair, Siddharth H and Wahlberg, Bo},
  booktitle={2023 IEEE Intelligent Vehicles Symposium (IV)},
  pages={1--8},
  year={2023},
  organization={IEEE}
}

@article{nair2024predictive,
  title={Predictive control for autonomous driving with uncertain, multimodal predictions},
  author={Nair, Siddharth H and Lee, Hotae and Joa, Eunhyek and Wang, Yan and Tseng, H Eric and Borrelli, Francesco},
  journal={IEEE transactions on control systems technology},
  volume={33},
  number={4},
  pages={1178--1192},
  year={2024},
  publisher={IEEE}
}

@article{muller2007dynamic,
  title={Dynamic time warping},
  author={M{\"u}ller, Meinard and others},
  journal={Information retrieval for music and motion},
  volume={69},
  pages={84},
  year={2007},
  publisher={Springer, Berlin, Heidelberg}
}

\end{document}